# Non-Iterative SLAM

Chen Wang, Junsong Yuan, and Lihua Xie
*School of Electrical and Electronic Engineering*
*Nanyang Technological University*
*50 Nanyang Avenue, 639798, Singapore*
*wang.chen@zoho.com;{jsyuan,elhxie}@ntu.edu.sg*

***Abstract*— The goal of this paper is to create a new framework for dense SLAM that is light enough for micro-robot systems based on depth camera and inertial sensor. Feature-based and direct methods are two mainstreams in visual SLAM. Both methods minimize photometric or reprojection error by iterative solutions, which are computationally expensive. To overcome this problem, we propose a non-iterative framework to reduce computational requirement. First, the attitude and heading reference system (AHRS) and axonometric projection are utilized to decouple the 6 Degree-of-Freedom (DoF) data, so that point clouds can be matched in independent spaces respectively. Second, based on single key-frame training, the matching process is carried out in frequency domain by Fourier transformation, which provides a closed-form non-iterative solution. In this manner, the time complexity is reduced to $\mathcal{O}(n \log n)$, where $n$ is the number of matched points in each frame. To the best of our knowledge, this method is the first non-iterative and online trainable approach for data association in visual SLAM. Compared with the state-of-the-arts, it runs at a faster speed and obtains 3-D maps with higher resolution yet still with comparable accuracy.**

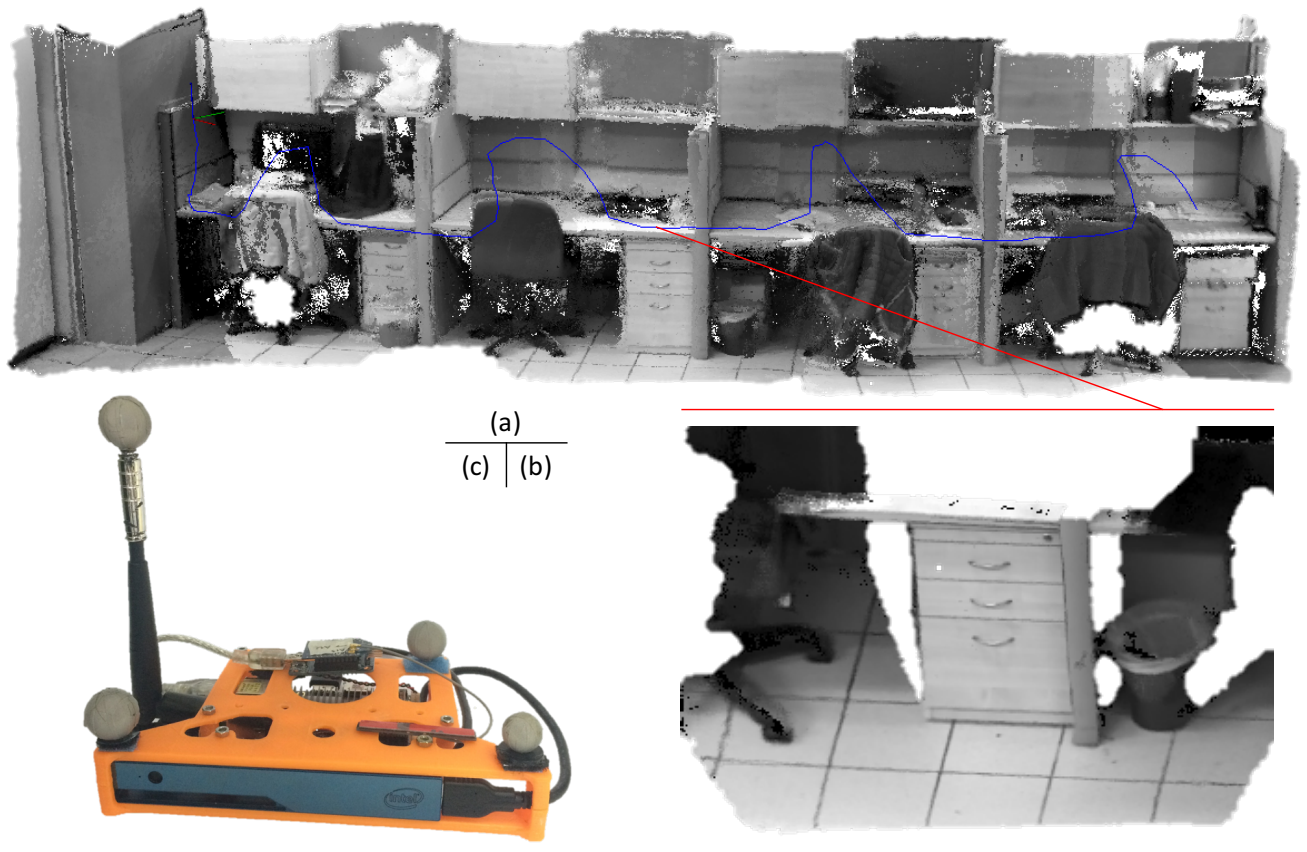

Fig. 1. The map (a) of a work place with resolution 0.005m is created by Non-Iterative SLAM (NI-SLAM) in real-time on an ultra-low power CPU with Scenario Design Power (SDP) of 2W shown in Table I. The map is shown without post-processing. (b) is an example of refined axonometric image. (c) is the integrated ultra-low power hardware platform, including an inertial sensor myAHRS+, Intel RealSense Robotic Development kit and a ZigBee module for receiving groundtruth from a motion capture system.

## I. Introduction

Vision based localization and mapping have received increasing attentions over the last decade. Despite the progress, one limitation of the existing methods is that the complexity of visual data association is too high for limited computational resources, especially when dense maps are required. In the SLAM problem, visual data association aims to relate the sensors' measurements with the landmarks in the map. It can be categorized as feature-based and direct methods.

Feature-based methods require a time consuming processing loop, including feature detection, extraction, matching, outliers rejection, and motion estimation. Generally, large number of features are required to be extracted in one image. To remove feature matching outliers, iterative methods like RANSAC are often used but they are time consuming. Moreover, in fitting a good motion model, iterative methods (e.g. Gauss-Newton) are needed to minimize the reprojection error [1], [2]. Therefore, the feature extraction process and the iterative solutions like RANSAC and gradient descent methods are the main computational burden.

For direct methods, iterative solutions also play an important role and have become the bottleneck. Motion model is estimated by the minimization of photometric error over all pixels [2]–[5]. Although feature extraction is no longer needed, it still needs heavy computation, since all the image points are involved in the minimization process based on iterative solutions such as Gauss-Newton method. On the other hand, Iterative Closest Point (ICP) is widely used for 3-D point clouds matching. Its time complexity is given by $\mathcal{O}(n^2)$ [6] which is very high to be carried out online ($n$ is the number of points to be matched).

To reduce the computational burden, it would be desirable to develop closed-form solutions for data association, which may require a new objective function. Interestingly, we find that if 3-D point clouds with 6 Degree-of-Freedom (DoF) can be decoupled into sub-spaces with lower DoF, the data association problem can be formulated as a classic regression problem with quadratic objective function, where a closed-form solution becomes feasible. Some works also proposed to fuse information from Inertial Measurement Unit (IMU) to improve performance. However, the information provided by IMU is still unable to significantly reduce the computation complexity of visual data association. To fully utilize the inertial information and reduce the complexity, we propose to decouple the 6 DoF data leveraging on attitude and heading reference system (AHRS) and axonometric projection. The key difference between IMU and AHRS is that the latter is equipped with an on-board processing unit providing superior reliable and accurate attitude and heading information, while the former just delivers sensor data to additional devices. In summary, the main contributions are:

- We propose a novel non-iterative framework for inertial-visual SLAM. It leverages on AHRS and depth camera and is light enough to be implemented with low-power processors.
- To enable faster data association, we propose a regression-based formulation and find a non-iterative solution with complexity $\mathcal{O}(n\log(n))$ by leveraging Fourier transform, where $n$ is the number of points. First, the 6 DoF point clouds are decoupled into 3 rotational DoF and 3 independent 1-D translational DoF. Second, instead of fitting a motion model iteratively, we estimate the camera motion by training the decoupled data online with a fast closed-form solution.
- A fast dense map refinement/fusion method is proposed based on a moving average. The missing information in one key-frame can be complemented by other point clouds with only the complexity $\mathcal{O}(n)$.

## II. Related Work

This paper focuses on visual data association and map creation without loop closure detection. Recent works on monocular and stereo camera with inertial fusion will also be reviewed, since many of those can be applied to process data from depth camera.

### A. Data Association

The idea of feature-based methods is that the incremental pose transformation can be estimated by well-matched features. Basically, there are two main approaches to find corresponding features across images. One is to extract features using local search techniques [2] and match them between the latest image and key-frame [7]. This method is suitable for images taken from nearby viewpoints. The other is to extract features independently and match them in a sequence of images based on similarity descriptors. This is suitable for large motion between two viewpoints. The matched features will be refined to remove outliers by RANSAC, then iterative gradient descent methods such as Gauss-Newton will be applied to find a motion transformation. The only information that conforms to the feature type can be used, which causes substantial loss of structural information.

In contrast with feature-based methods, direct methods match all image points directly, resulting in a denser map which provides substantially more information about the geometry of environment. Based on the minimization of photometric error, direct methods have been developed for monocular [4], stereo [8], and depth [3] cameras. Since there is no descriptor, only the local search techniques can be used to find the corresponding pixels. Moreover, since more image points are involved to fit a pose transformation model by an iterative solution, the time complexity is still very high for low-power systems.

Many dense mapping systems resort to ICP to align point clouds with respect to photometric or geometric constraints. Compared with feature-based and direct methods, works like [9]–[11] are able to create dense maps which are crucial for higher level applications, such as object detection and scene understanding. KinectFusion [9] is one of the most famous methods. The incremental pose transformation is obtained by tracking the live depth frame relative to the global model using a coarse-to-fine ICP with geometric constraints. However, the ICP-based methods require powerful GPU to process large number of data, they are not applicable to systems with low computational power.

### B. Inertial Fusion

Inertial sensors provide an additional constraint for the pose transformation estimation which can help speed up the visual SLAM systems [12]. Loosely-coupled methods [13] fuse the pose estimates from SLAM and IMU independently. While tightly-coupled methods which estimate the states jointly give a better performance in terms of accuracy, but have an additional complexity due to the involvement of a non-linear optimization process [14]. In recent years, IMU fusion has been proposed for feature-based-monocular [15], feature-based-stereo [14], direct-stereo [13], and direct-monocular [12] systems.

## III. Iterative SLAM

The Iterative SLAM in this paper is defined as those SLAM algorithms that need iterative solutions to find data association. The ICP-based methods such as [9]–[11] are the most representative examples of Iterative SLAM. Let $\{\mathcal{M},\mathcal{S}\}$ be two finite size point sets, the objective of ICP is to find a transformation $\mathcal{T}(\,\cdot\,)$ to be applied to the point set $\mathcal{M}$, such that the distance $D(\,\cdot\,)$ between $\mathcal{T}(\mathcal{M})$ and $\mathcal{S}$ is minimized.

$$D(\mathcal{T}(\mathcal{M}),\mathcal{S}) := \sum_{\mathbf{m}\in\mathcal{T}(\mathcal{M})}\sum_{\mathbf{s}\in\mathcal{S}}\rho(\mathbf{m}-\mathbf{s}), \qquad (1)$$

where $\rho(\,\cdot\,)$ is a general objective function, and can be the square of the Euclidean distance for the simplest case. Again, the complexity of solving the objective (1) by iterative solutions is too high for low-power systems and no optimal solution is guaranteed.

Those methods that minimize the reprojection or photometric error to estimate the pose transformation are also the examples of Iterative SLAM. Assume that the camera model $\pi(\,\cdot\,)$ projects 3-D point $p_i$ to image point $u_i$, so that $u_i = \pi(E_{CW}p_i)$, where $E_{CW}$ is the transform matrix from the world to camera frame. Incremental poses $E'_{CW}$ are obtained by left-multiplying $E_{CW}$ with a vector of 6 DoF based on exponential mapping, i.e., $E'_{CW} = exp(\mu)E_{CW}$.

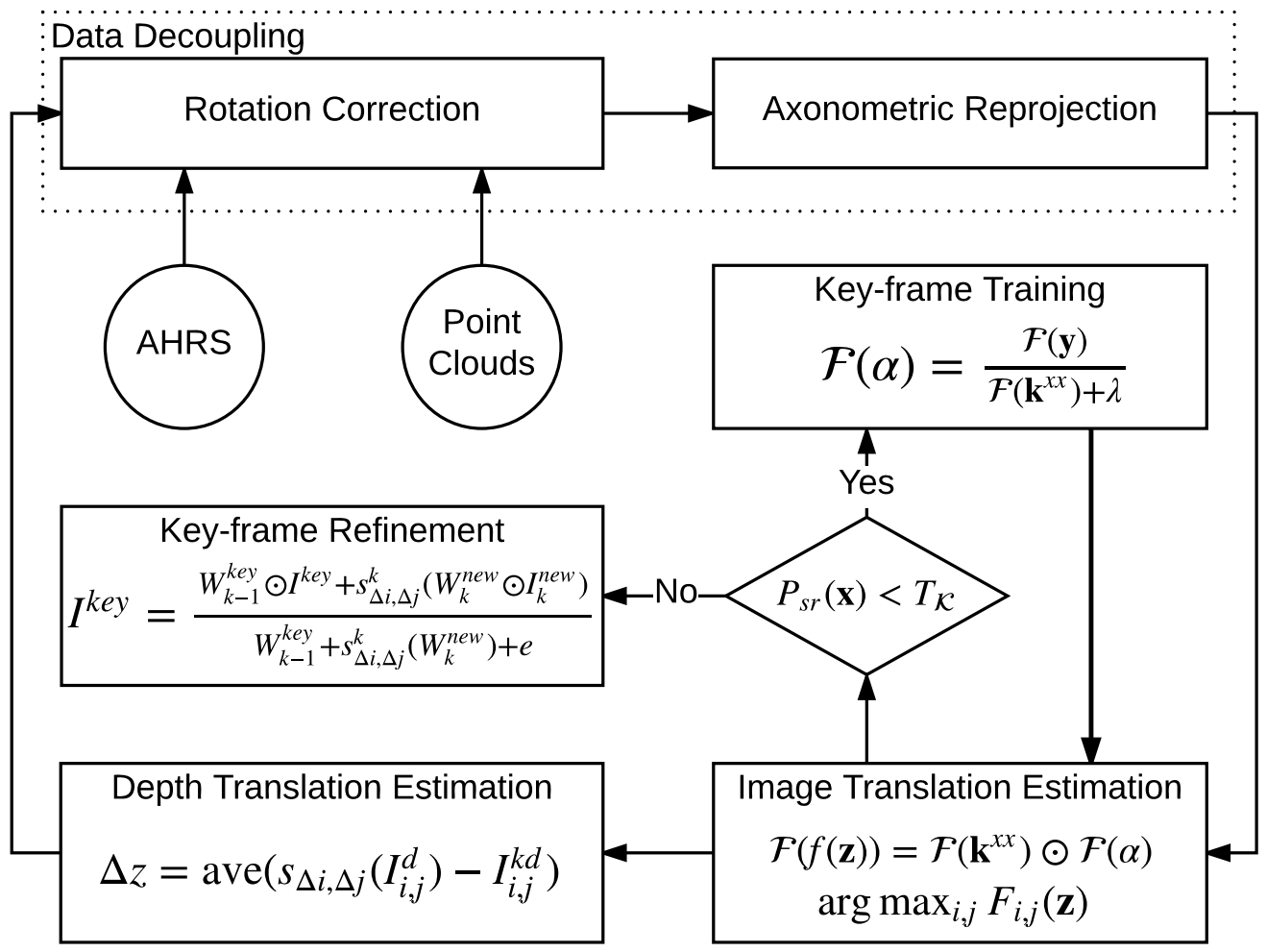

Fig. 2. The proposed framework of Non-Iterative SLAM. Only when the maximum response value is lower than a threshold, a new key-frame will be inserted into the map, or the previous key-frame will be refined.

In feature-based methods, feature points $\mathbf{u}_i$ in one image are reprojected to another image $\mathbf{u}'_i$, so that the incremental pose transformation vector $\mu$ can be estimated by minimizing the reprojection error $\mathcal{R}(\mu)$.

$$\mathcal{R}(\mu) = \sum_i \rho(\mathbf{u}'_i - \pi_p(E'_{CW}\pi_p^{-1}(\mathbf{u}_i, E_{CW}))), \quad (2)$$

where $\rho$ is a general objective function. In [1], $\rho$ is a Tukey's biweight function; whereas in [2], [16], $\rho$ is an $L_2$-norm function. For monocular camera, an initial process to obtain the depths of points is needed [1], [2], [7], while for stereo/RGB-D camera, this process can be skipped.

In direct methods, new images are aligned with several key frames $I_r$ by minimizing the photometric error $\mathcal{H}(\mu)$ given in (3). Similar expressions can be found in [2], [4], [5].

$$\mathcal{H}(\mu) = \sum_{i,r} \rho(I_r(\mathbf{u}'_i) - I(\pi_p(E'_{C_rW}\pi_p^{-1}(\mathbf{u}_i, E_{CW})))), \quad (3)$$

where $\rho(\cdot)$ has the same meaning with (2). Being non-robustness to illumination changes is one of the potential problems of matching pixels [5]. Since (2) and (3) are highly non-linear, they can only be solved via an iterative method. Unfortunately, iterative methods are sensitive to initialization and cannot guarantee the global optima.

## IV. Non-Iterative SLAM

To avoid iterative solutions and reduce computational burden, we propose a novel framework for Non-Iterative SLAM shown in Fig. 2. First, in Section IV-A, the 6 DoF point cloud is decoupled and reprojected to axonometric images with only 1 DoF. Section IV-B demonstrates that the 1 DoF data can be matched by a closed-form non-iterative solution. The decoupled translations are estimated in Section IV-C and IV-D respectively. Finally, refinements of the key-frames are presented in Section IV-E. It should be noted that the term 'non-iterative' is only used for data association, while the process of mapping and localization are still iterative.

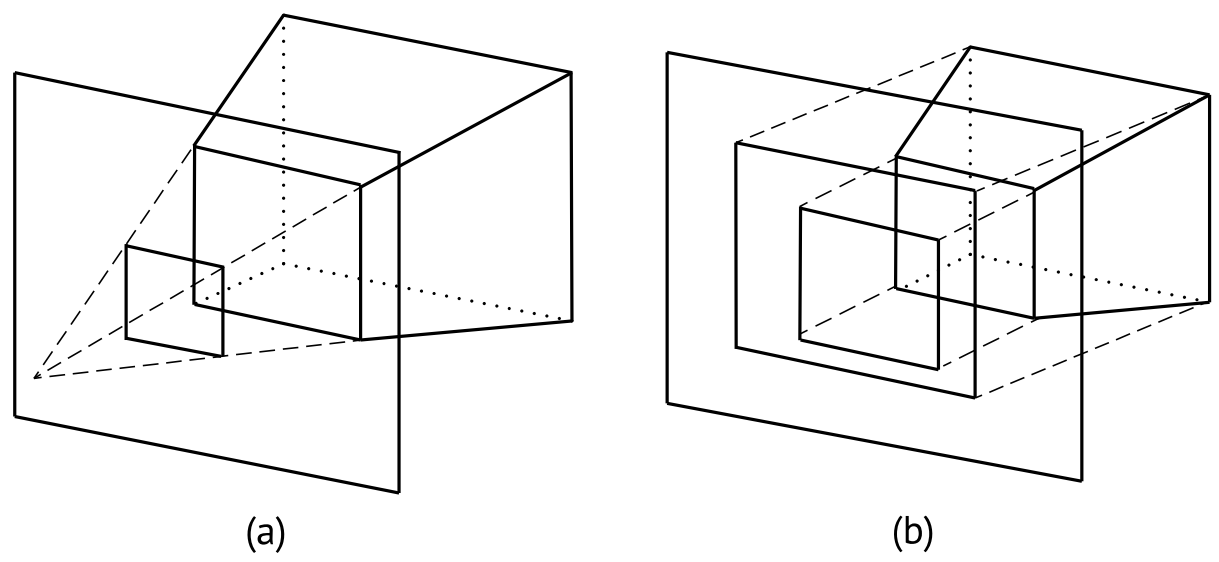

Fig. 3. Differences between perspective and axonometric projections. A quadrangular frustum pyramid can be a rectangle in perspective image (a) when the vertex coincides with the principle point. While for axonometric image (b), the size ratio of the two rectangles will not change. This property will be used in the procedure of data association.

### A. Point Clouds Reprojection

*1) 6 DoF to 3 DoF:* A point cloud is a set of data points with 6 DoF $\mu(x, y, z, \alpha, \beta, \phi)$ in a three-dimensional coordinate system. To decouple the rotational and translational DoF, we utilize the attitude information from AHRS directly. An AHRS consists of sensors on three axes that provide attitude information for aircraft, including roll, pitch, and yaw. In this sense, the 6 DoF data $\mu(x, y, z, \alpha, \beta, \phi)$ can be reduced to 3 DoF $\mu(x, y, z)$ easily.

*2) 3 DoF to 2 DoF:* The basic idea for decoupling the 3 translational DoF $\mu(x, y, z)$ is that the geometry properties in the 3 separate axes must be kept. Therefore, we propose to apply axonometric projection on the rectified point clouds to get axonometric color and depth images. Different from the perspective projection which projects 3-D points on the principle point, the axonometric projection projects points onto the axonometric plane shown in Fig. 3. Since the distance ratio in the axonometric plane between any pair of image points will not change, the 3-D translation can be estimated in the 2-D axonometric image plane followed by 1-D translation estimation in the depth direction.

Since point clouds are generated by pinhole cameras, when they are reprojected to the axonometric plane, some black holes may appear. The reason is that some invisible points for pinhole cameras are also projected to the axonometric plane. It will be shown that these black holes do not affect the translation estimation and will be filled by the refinement process for the key-frames in Section IV-E. Fig. 4 shows the procedure that a 6 DoF point cloud is rectified by AHRS then reprojected to axonometric images.

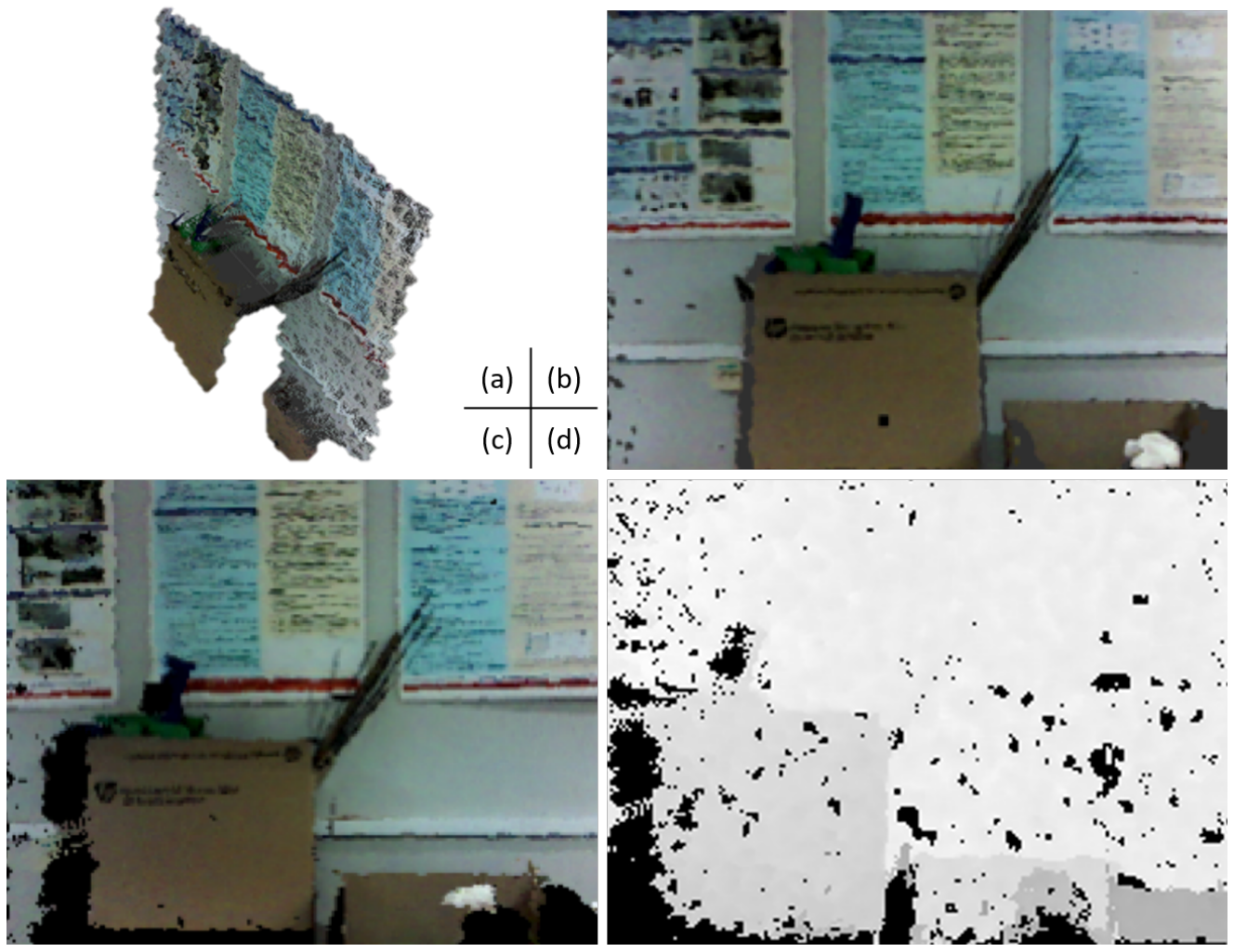

Fig. 4. A point cloud (a) captured by Intel RealSense is rectified by AHRS shown in (b) which is a perspective image. Then it is reprojected to an axonometric color image (c) and axonometric depth image (d). Some black holes can be seen in (c) and (d), since some invisible points in (b) are also projected to the axonometric image plane.

*3) 2 DoF to 1 DoF:* The core idea is that the translation of a 2-D axonometric image can be treated as 1-D translation of an expanded vector, which is illustrated in Fig. 5. The data decoupling procedure from 6 DoF to 1 DoF does not lose any structural information, since the original 6 DoF data can be restored without additional information.

### B. Translation Pattern Training

The fundamental idea is that we want to learn the pattern of an expanded key axonometric image so that the translation of a new image can be predicted directly based on the assumption that there is enough overlap between the key and the new frames. Assume that the $N$ by $M$ key axonometric image is denoted by a column vector $\mathbf{x} \in \mathbb{R}^n$, where $n = N \times M$. Hence the training samples can be defined as $X = [\mathbf{x}_0, \mathbf{x}_1, \cdots, \mathbf{x}_{n-1}]^T$ where $\mathbf{x}_i$ is obtained from cyclic shifted $\mathbf{x}$ by $i$ pixels. Basically, $X$ is a circulant matrix and can be defined as $X = C(\mathbf{x})$, since it can be generated by its first row $\mathbf{x}^T$. The training objective is to find a regression function $y_i = f(\mathbf{x}_i)$ where each sample $\mathbf{x}_i$ is with a target $y_i$, so that when a test sample $\mathbf{z}$ is given, $f(\mathbf{z})$ will be corresponding to a translation-related target. Intuitively, if $y_i = i$, the translation of the test samples can be predicted directly. However, since only the zero-shift sample $x_0$ is concerned, the label $y_0$ is set as 1, while all the others are 0. In this sense, all the non-zero-shifts are considered as negative samples, which makes the regression function more distinctive.

*1) Linear Case:* A linear regression function is defined as $f(\mathbf{x}) = \mathbf{w}^T\mathbf{x},\ \mathbf{w} \in \mathbb{R}^n$. Given the training samples $X$, the objective is to find the coefficient vector $\mathbf{w}$ by minimizing

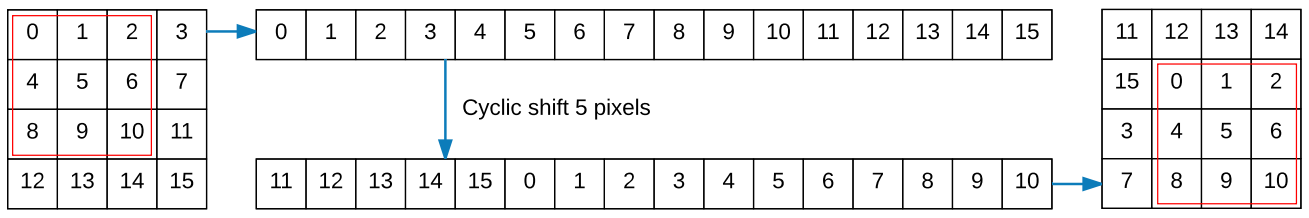

Fig. 5. The example of the data decoupling from 2 DoF to 1 DoF. The axonometric images are expanded into a 1-D vector, and cyclic shift of the 1-D vector can be mapped to a unique 2-D translation in the original image. Since it is a one-to-one mapping, the translation estimation in the 2-D image plane can be replaced by a 1-D vector. The disadvantage of cyclic shift is the border effect, where the pixels on one side of the original image may be relocated to the other side of the new image. However, these elements contribute nothing for finding the overlap of the two images (bounded by the red rectangle), so that they can just be ignored.

the squared error over the regression function and the target.

$$\mathbf{w}^\star = \arg\min_{\mathbf{w}} \sum_{i=1}^{n} (f(\mathbf{x}_i) - y_i)^2 + \lambda \left\|\mathbf{w}\right\|^2, \tag{4}$$

where $\lambda$ is a regularization parameter to prevent overfitting. This is a ridge regression problem and can be solved by setting its first derivative to zero. Define $\mathbf{y} = [y_0,\ y_1,\ \cdots,\ y_{n-1}]^T$, a closed-form solution (5) can be found in complex domain [17]:

$$\mathbf{w}^\star = \left(X^H X + \lambda I\right)^{-1} X^H \mathbf{y}, \tag{5}$$

where $H$ denotes the conjugate transpose. For a $480 \times 360$ axonometric image, where $n = 172800$, it requires to compute the inversion of $172800 \times 172800$ matrix $(X^T X + \lambda I)$, which is impossible to be carried out in real-time. However, the interesting thing is that, different from a traditional machine learning problem, $X = C(\mathbf{x})$ is a circulant matrix. This amazing property makes the solution (5) easy to be obtained due to the following lemma.

***Lemma** 1 (J. F. Henriques [18]):* If $X$ is a circulant matrix, the solution (5) can be converted into frequency domain:

$$\mathcal{F}(\mathbf{w}^\star) = \frac{\mathcal{F}^*(\mathbf{x}) \odot \mathcal{F}(\mathbf{y})}{\mathcal{F}^*(\mathbf{x}) \odot \mathcal{F}(\mathbf{x}) + \lambda}, \tag{6}$$

where $\mathcal{F}(\,\cdot\,)$ is the discrete Fourier transform and the superscript operator $^*$ is the complex conjugate, $\odot$ and $-\!\!\!\!\div$ are the element-wise multiplication and division respectively.

Except for the Fourier transform, all the operations in (6) are element-wise. Therefore, the complexity is dominated by the fast Fourier transform (FFT) $\mathcal{O}(n\log(n))$ where $n$ is the number of points in the axonometric image. While the complexity of the original solution (5) is dominated by a matrix inversion, whose complexity has a lower and upper bounds, given by matrix multiplication $\mathcal{O}(n^2\log(n))$ and Gauss-Jordan elimination method $\mathcal{O}(n^3)$. For a $480 \times 360$ image, where $n = 172800$, the complexity ratio $r \in [\mathcal{O}(n^2\log(n))/\mathcal{O}(n\log(n)),\ \mathcal{O}(n^3)/\mathcal{O}(n\log(n))] = [172800,\ 2480000000]$, which implies that lots of running time can be saved if the problem (4) is solved by (6).

*2) Non-linear Case:* Data samples may be linearly separable in a high dimension space although they are not in the original space. Suppose $\phi(\cdot)$ is a high dimension feature space, such that $\phi : \mathbb{R}^n \rightarrow \mathbb{R}^d$ where $d \gg n$, a kernel $k$ is the inner product of the feature mapping:

$$k(\mathbf{x}, \mathbf{z}) = \phi(\mathbf{x})^T \phi(\mathbf{z}), \tag{7}$$

where $\mathbf{z}$ is a test sample. The solution $\mathbf{w} \in \mathbb{R}^d$ is expressed as a linear combination of training data $\mathbf{x}_i$ in the feature space:

$$\mathbf{w} = \sum_{i=0}^{n-1} \alpha_i \phi(\mathbf{x}_i). \tag{8}$$

The regression function becomes

$$f(\mathbf{z}) = \mathbf{w}^T \phi(\mathbf{z}) = \sum_{i=0}^{n-1} \alpha_i k(\mathbf{x}_i, \mathbf{z}). \tag{9}$$

Then minimizing the original objective function (4) is equivalent to finding the combination coefficient $\alpha = [\alpha_0, \alpha_1, \cdots, \alpha_{n-1}]^T$. Given the training data $X$, the solution of (4) becomes [17],

$$\alpha = (K + \lambda I)^{-1} \mathbf{y}, \tag{10}$$

where $K$ is the kernel matrix with each element $k_{ij} = k(\mathbf{x}_i, \mathbf{x}_j)$. The dimension of $\alpha$ depends on the number of samples that is the length of $\mathbf{x}$. Fortunately, the kernel matrix $K$ is circulant when $k(\mathbf{x}_i, \mathbf{x}_j)$ is a Gaussian kernel [18]:

$$k(\mathbf{x}_i, \mathbf{x}_j) = -\frac{1}{2\pi\sigma^2} e^{-\frac{\|x_i - x_j\|^2}{2\sigma^2}}. \tag{11}$$

Therefore, (10) can be calculated in frequency domain with complexity $\mathcal{O}(n \log(n))$:

$$\mathcal{F}(\alpha) = \frac{\mathcal{F}(\mathbf{y})}{\mathcal{F}(\mathbf{k}^{\mathbf{xx}}) + \lambda}, \tag{12}$$

where $\mathbf{k}^{\mathbf{xx}}$ is the first row of the kernel matrix $K = C(\mathbf{k}^{\mathbf{xx}})$. To guarantee the robustness, all the circular shifts of a sample $\mathbf{z}$ are tested. Define the kernel matrix $K^{\mathbf{zx}}$ where each element $K_{ij} = k(\mathbf{z}_i, \mathbf{x}_j)$ and $\mathbf{z}_i$ is the $i_{th}$ row of the circulant matrix $C(\mathbf{z})$, from (9), we have

$$\mathbf{f}(\mathbf{z}) = K^{\mathbf{zx}} \alpha, \tag{13}$$

where $\mathbf{f}(\mathbf{z}) = [f(\mathbf{z}_0), f(\mathbf{z}_1), \cdots, f(\mathbf{z}_{n-1})]^T$. Since $K^{\mathbf{zx}}$ is a circulant matrix, again we have $K^{\mathbf{xz}} = C(\mathbf{k}^{\mathbf{zx}})$ where $\mathbf{k}^{\mathbf{zx}}$ is the first row of $K^{\mathbf{zx}}$. Therefore, the response of the circular shifted samples can be found in the frequency domain:

$$\mathcal{F}(\mathbf{f}(\mathbf{z})) = \mathcal{F}(\mathbf{k}^{\mathbf{zx}}) \odot \mathcal{F}(\alpha). \tag{14}$$

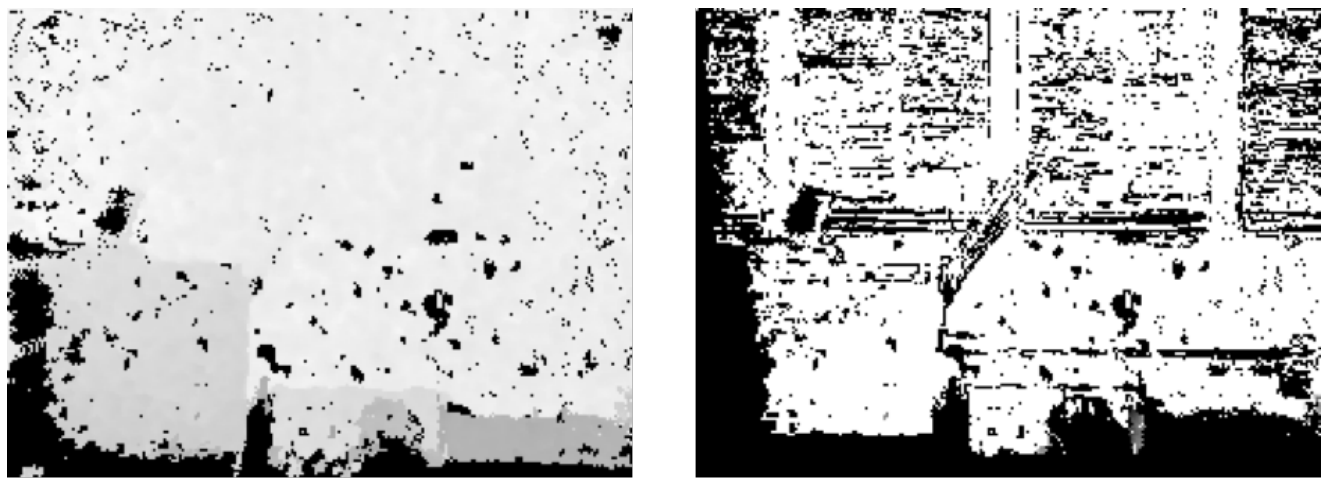

(a) (b)

Fig. 6. The left image (a) is an axonometric depth image captured from real-time data. The corresponding well-matched points defined in (18) are shown in the right image (b) by pixel intensities. The higher the brightness, the more confidence the matches have.

### C. Image Translation Estimation

The response vector $\mathbf{f}(\mathbf{z})$ is reshaped back to the original axonometric plane which is denoted as response matrix $\mathbf{F}(\mathbf{z})_{N \times M}$. Since only the zero-shift label $y_0$ is set as 1, the estimated translation of test sample $\mathbf{z}$ should be corresponding to the location of the maximum value in the response matrix $\mathbf{F}(\mathbf{z})$. Let the estimated translation of the axonometric image be denoted as $(\Delta i, \Delta j)$ with pixel unit. Then the estimated translation on the axonometric plane $(\Delta x, \Delta y)$ can be denoted as element-wise multiplication:

$$(\Delta x, \Delta y) = (r_x, r_y) \odot (\Delta i, \Delta j), \tag{15}$$

where $r_x$ and $r_y$ are the image resolutions in $x$ and $y$ directions respectively.

As the camera moves, the overlap between the key and new frames decreases, resulting in a weak peak strength. A measurement of peak strength is called the Peak to Sidelobe Ratio (PSR). In the definition of (16), the output $\mathbf{F}_{i,j}$ is split into the peak which is the maximum value and the sidelobe which is the rest of pixels excluding the peak.

$$PSR = \frac{\max \mathbf{F}_{i,j}(\mathbf{z}) - \mu_s}{\sigma_s}, \tag{16}$$

where $\mu_s$ and $\sigma_s$ are the mean and standard deviation of the sidelobe. PSR is a similarity measurement of two point clouds and is considered as a trigger to insert a new key-frame into the map. The condition is

$$PSR < T_r, \tag{17}$$

where $T_r$ is a predefined threshold. PSR criterion (17) is not only able to control the minimum confidence of each matching, but also save computational time, especially when the camera is kept still or moving slowly since there is no new training data required.

### D. Depth Translation Estimation

The translation $\Delta z$ in depth direction is estimated in (18). To be robust, it averages the differences of the well-matched depth pixels $I^d_{i,j}$ defined in (18). Fig. 6 shows an example

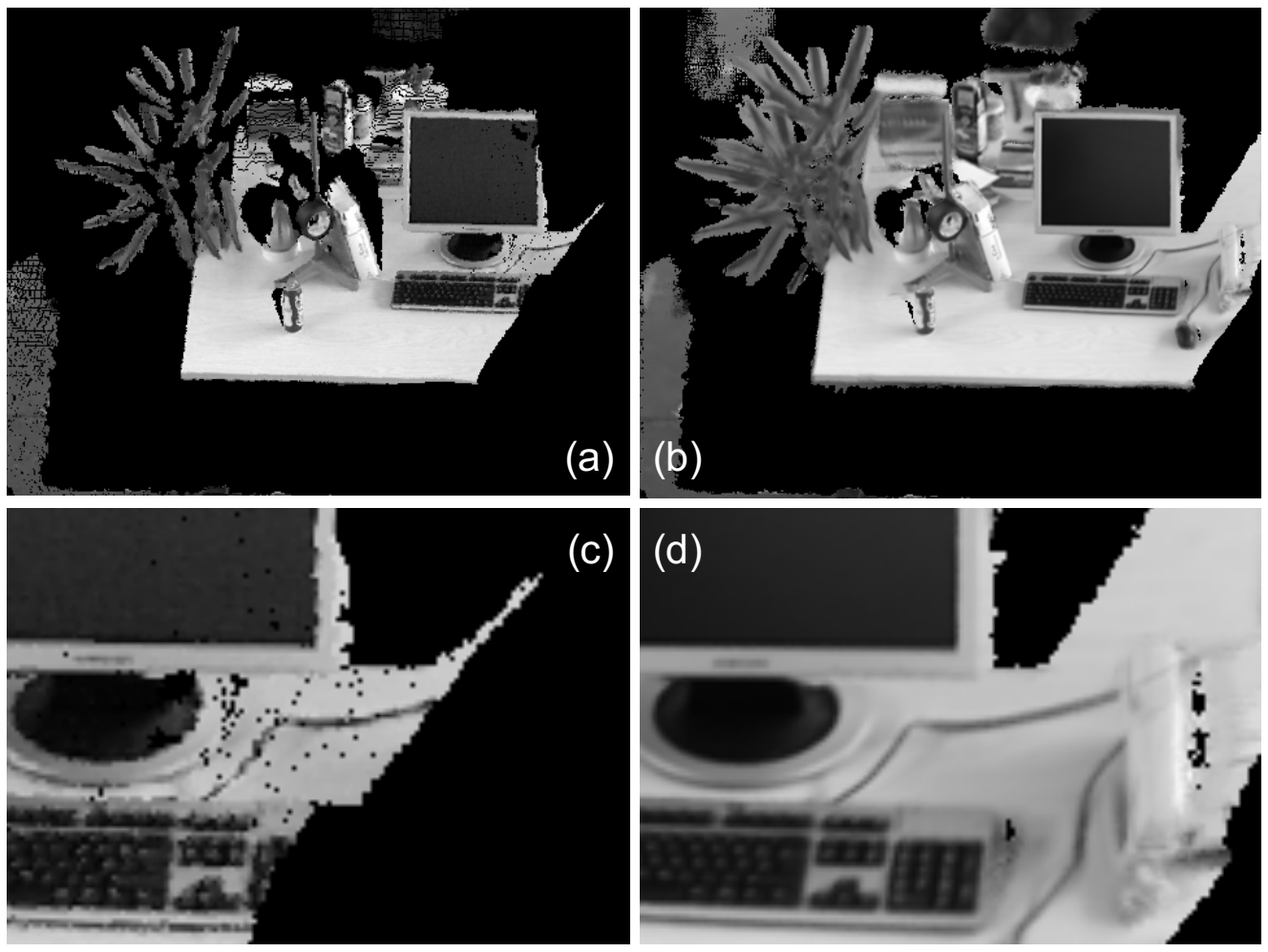

Fig. 7. (a) is a new key cloud. (b) is the refined key cloud during run time. Tests show that the "black holes" in the original axonometric image can be filled by subsequent matched images based on Equation (19). (c) and (d) are the same part of new and refined key point cloud in (a) and (b) respectively. Note that the keyboard on the desk is nearly perfectly complemented. The moving average operation will not make it blurred, if the image translation is estimated correctly.

of axonometric depth image and the corresponding well-matched points.

$$\Delta z = \text{ave}\left(s_{\Delta i,\Delta j}(I^{d}_{i,j}) - I^{kd}_{i,j}\right), \tag{18}$$

where $(i,j) \in \left\{(i,j)|\rho\left(s_{\Delta i,\Delta j}(I^{c}_{i,j}) - I^{kc}_{i,j}\right) < T_c\right\}$ and $\rho(\cdot)$ is a general objective function ($L_1$-norm in the tests). $(\Delta i, \Delta j)$ is the estimated image translation in (15) and $s_{\Delta i,\Delta j}(\cdot)$ is the shift of an image by $(\Delta i, \Delta j)$ pixels. The superscript $d, kd, c, kc$ of image $I$ are the depth, key depth, color and key color images respectively.

The advantage of (18) is that it only requires the average differences of the matched depth pixels, which is extremely fast to compute and all the well-matched points are able to contribute to the estimation. This makes it robust to the depth noises. Therefore, the translation estimation $\Delta p = [\Delta x, \Delta y, \Delta z]^T$ is obtained based on the decoupled translation in the axonometric plane and depth direction. Next, a refinement of key point clouds will be presented.

### *E. Refinement of Key Point Clouds*

When the camera is kept still or moving slowly, the overlap between the new and key-frame is large enough and criterion (17) is not satisfied. The $k_{th}$ new frame $I^{new}_k$ can be used to refine the key frame $I^{key}$. A moving average defined in (19) is applied to refine both the color and depth information.

$$I^{key} = \frac{W^{key}_{k-1} \odot I^{key} + s^{k}_{\Delta i,\Delta j}(W^{new}_k \odot I^{new}_k)}{W^{key}_{k-1} + s^{k}_{\Delta i,\Delta j}(W^{new}_k) + e}, \tag{19a}$$

$$W^{key}_k = W^{key}_{k-1} + W^{new}_{k-1}, \tag{19b}$$

where $e$ is a small term (set as $1e^{-7}$) to prevent division by 0. The size of the weight matrix $W$ is the same as the image $I$ and each pixel presents the weight of that pixel to be fused. $W^{new}_k(i,j) = \{0,1\}$ where 1 or 0 indicates whether the pixel $(i,j)$ can be seen in the original point cloud or not. Hence the matrix $W^{new}_k$ can be obtained in parallel with the procedure of reprojection. $W^{key}_0$ is initialized as $W^{new}$ when the point cloud $I^{new}$ is inserted as a key-frame $I^{key}$. Fig. 7 shows the example of a new and refined key-frame. For axonometric depth image, the only difference is that the term $I^{new}_k$ in (19a) will be replaced by $I^{new}_k - \Delta z_k$, where $\Delta z_k$ is the estimated depth translation at time step $k$.

## V. Evaluation

We evaluate our algorithm and compare it with the state-of-the-art methods Volumetric Fusion [10] and RGB-D SLAM [19], [20] in terms of computational speed, map resolution, and accuracy of estimated trajectory. All the three indexes are evaluated on the widely used RGB-D benchmark [21] that provides synchronized ground truth from a motion capture system, which is also used to produce AHRS data.

We evaluate multi-runs over 10 different datasets each of which contains normal translational and rotational movements by a hand-held camera. Besides, the datasets fr3/nst and fr3/stf, fr3/stn contain **n**onstructual/**s**tructual environment at **n**ear/**f**ar viewpoint; fr3/sitting-static and fr3/sitting-xyz contain dynamic objects in the field of view.

Experiments show that hardware platform has a significant impact on the performance. Hence, we compare the platforms together with the corresponding update rate and map resolution in Table I (NI-SLAM for evaluation). Typically, higher resolution will result in slower update rate. However, we still achieve the fastest update rate with the highest map resolution using the lowest-power CPU without any GPU devices.

More details about the distribution of update frequency over all the 10 datasets can be found in the box plot of Fig. 8 which shows that all the median update frequencies are above 50Hz. The corresponding statistic performance of the accuracy is shown in Table II. We use the absolute trajectory root-mean-square error metric (RMSE) to evaluate our system. The comparison with the state-of-the-art algorithms can be found in Table III where '−' indicates that the data can not be found or that algorithm is not able to produce an estimate on the dataset. The performances of [10], [19], [20] are cited from their papers. Table I and III indicate that we achieve the fastest running speed with the highest map resolution and comparable accuracy.

## VI. On-the-fly Testing

A low-power platform is chosen for the on-the-fly testing. It features a coin-sized inertial sensor myAHRS+, an Intel RealSense R200 camera, and a credit card-sized processing board. Running at 1.44GHz with 2G RAM, this platform

TABLE I

THE COMPARISON OF PLATFORM FOR SYSTEM EVALUATION. THE UPDATE RATE AND MAP RESOLUTION ARE SHOWN IN THE LAST COLUMN.

| Methods | CPU | RAM | GPU | Update Rate/Resolution |
|---|---|---|---|---|
| NI-SLAM (evaluation) | i7-5550U 2.0GHz Dual Core | 8Gb | None | 50Hz/0.005m |
| NI-SLAM (micro-robots) | Atom x5-Z8350 1.44GHz | 2Gb | None | 15Hz/0.005m |
| Volumetric Fusion [10] | i7-3960X 3.3GHz Hexa Core | 16Gb | nVidia GTX680 | 30Hz/0.010m |
| RGB-D SLAM [19], [20] | i7-***** 3.4GHz Quad Core | 8Gb | nVidia GTX570 | 30Hz/0.100m |

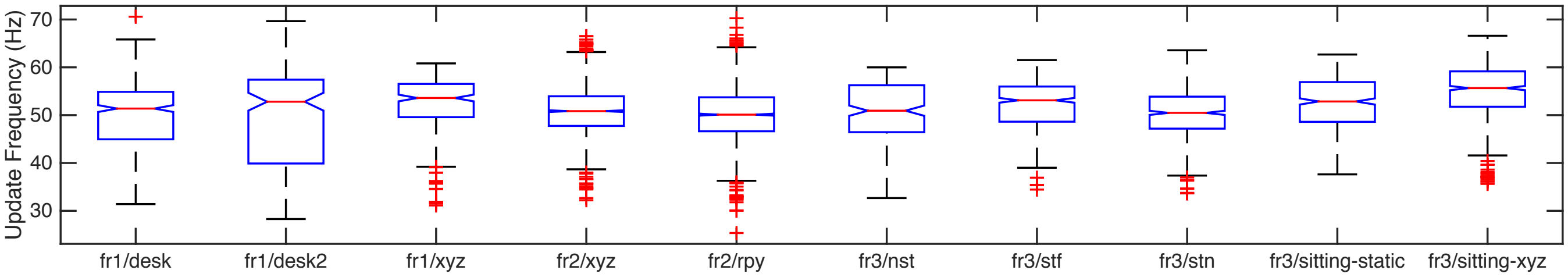

Fig. 8. Update rates are all above 50Hz with the axonometric image size 360 × 480. The training time of key-frames is also considered in each update.

TABLE II

THE ACCURACY PERFORMANCE ON THE BENCHMARK [21]. THE MEAN TRANSLATIONAL AND ANGULAR VELOCITY ARE GIVEN IN $\bar{v}$ AND $\bar{\omega}$ RESPECTIVELY. 'DYNAMIC' MEANS DYNAMIC OBJECTS CAN BE SEEN IN THE FIELD OF VIEW.

| Dataset | RMSE (m) | Mean (m) | Median (m) | Std. (m) | Resolution (m) | $\bar{v}$ (m/s) | $\bar{\omega}$ (°/s) | Dynamic |
|---|---|---|---|---|---|---|---|---|
| fr1/desk | 0.0252 | 0.0236 | 0.0216 | 0.0087 | 0.0020 | 0.413 | 23.32 | No |
| fr1/desk2 | 0.0598 | 0.0563 | 0.0530 | 0.0200 | 0.0025 | 0.426 | 29.31 | No |
| fr1/xyz | 0.0112 | 0.0097 | 0.0083 | 0.0055 | 0.0035 | 0.244 | 8.92 | No |
| fr2/xyz | 0.0121 | 0.0108 | 0.0102 | 0.0054 | 0.0030 | 0.058 | 1.72 | No |
| fr2/rpy | 0.0269 | 0.0225 | 0.0202 | 0.0147 | 0.0050 | 0.014 | 5.77 | No |
| fr3/nst | 0.0171 | 0.0157 | 0.0153 | 0.0068 | 0.0060 | 0.299 | 2.89 | No |
| fr3/stf | 0.0137 | 0.0127 | 0.0122 | 0.0051 | 0.0060 | 0.193 | 4.32 | No |
| fr3/stn | 0.0193 | 0.0183 | 0.0174 | 0.0061 | 0.0040 | 0.141 | 7.68 | No |
| fr3/siting-static | 0.0097 | 0.0086 | 0.0077 | 0.0045 | 0.0060 | 0.011 | 1.70 | Yes |
| fr3/siting-xyz | 0.0514 | 0.0439 | 0.0382 | 0.0267 | 0.0060 | 0.132 | 3.56 | Yes |

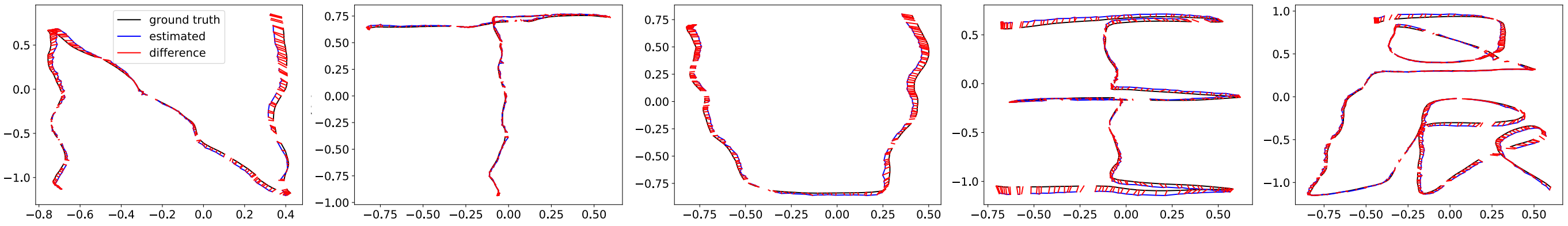

Fig. 9. The top view of the trajectory estimation on the low-power platform.

shown in Fig. 1 (c) is very difficult for most of the state-of-the-art algorithms to run in real-time. The efficiency performance and more details about this platform can be found in Table I (NI-SLAM for micro-robots). A work place map created in real-time on this low-power platform is shown in Fig. 1 (a) without any post-processing. The map is updated in 15Hz with resolution 0.005m. Hence, much details of the environments can be preserved. We also run the proposed framework 5 times in a room equipped with a motion capture system. Fig. 9 shows the estimated trajectory from the top of view compared with ground truth. The corresponding update rates and accuracy performance are shown in Fig. 10 and Table IV. It can be seen that, although limited by the low quality of point clouds (RealSense camera produces more noises than Kinect), our proposed framework can still work well on the low-power platform.

TABLE III

COMPARISON OF ACCURACY ON THE BENCHMARK [21]. (UNIT: m)

| Dataset | NI-SLAM | Vol.-Fusion [10] | RGB-D [19], [20] |
|---|---|---|---|
| fr1/desk | **0.025** | 0.037 | 0.026 |
| fr1/desk2 | **0.060** | 0.071 | 0.102 |
| fr1/xyz | **0.011** | 0.017 | 0.021 |
| fr2/xyz | 0.012 | 0.029 | **0.008** |
| fr2/rpy | **0.027** | - | - |
| fr3/nst | **0.017** | 0.031 | 0.018 |
| fr3/stf | **0.014** | - | - |
| fr3/stn | **0.019** | - | - |
| fr3/sitting-static | **0.010** | - | - |
| fr3/sitting-xyz | **0.051** | - | - |

TABLE IV

PERFORMANCE OF ACCURACY ON LOW-POWER PLATFORM. (UNIT: m)

| Tests | RMSE | Mean | Median | Std. | Resolution |
|---|---|---|---|---|---|
| First | 0.0318 | 0.0280 | 0.0268 | 0.0148 | 0.005 |
| Second | 0.0140 | 0.0129 | 0.0128 | 0.0054 | 0.005 |
| Third | 0.0327 | 0.0307 | 0.0292 | 0.0111 | 0.005 |
| Forth | 0.0385 | 0.0345 | 0.0301 | 0.0171 | 0.005 |
| Fifth | 0.0313 | 0.0282 | 0.0275 | 0.0135 | 0.005 |

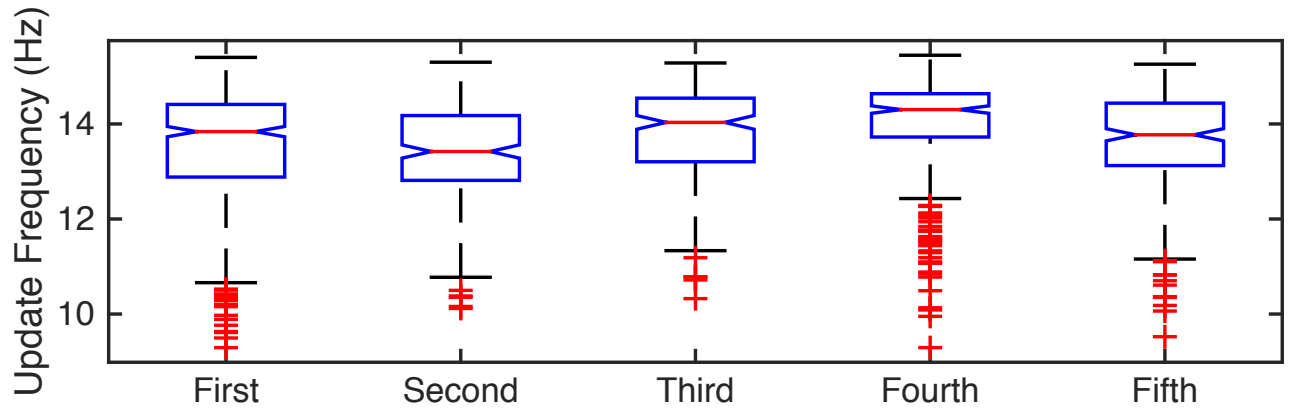

Fig. 10. Update rates on the low-power platform. The training time of key frames is also considered in each update.

## VII. CONCLUSION

A novel non-iterative framework for inertial-visual SLAM using depth camera is proposed in this paper. First, different from existing methods, our algorithm fuses attitude information from AHRS and decouples the original 6 DoF data by axonometric projection. Second, by leveraging the property of circulant matrix and also Fourier transform, we find a non-iterative solution for visual data association, which significantly decreases the computational burden and makes real-time dense SLAM feasible for micro-robots. Last, a fast map refinement/fusion method is designed based on moving average on the rectified point clouds. To the best of our knowledge, the proposed framework is the first non-iterative and online trainable solution for dense SLAM and achieves a faster speed with higher map resolution and comparable accuracy compared with the state-of-the-art algorithms.